\documentclass[times,twocolumn,final]{elsarticle}

\usepackage{medima}
\usepackage{framed,multirow}

\usepackage{amssymb}
\usepackage{latexsym}

\usepackage{url}
\usepackage{xcolor}

\usepackage{todonotes}
\usepackage[colorlinks=true]{hyperref}
\usepackage{mathtools}
\usepackage{amsmath}
\usepackage{amsfonts}
\DeclarePairedDelimiter\ceil{\lceil}{\rceil}

\usepackage{latexsym,graphicx,array,amsfonts}

\usepackage{multirow}
\usepackage{algorithm}
\usepackage[noend]{algpseudocode}
\usepackage{array}
\newcolumntype{L}[1]{>{\raggedright\let\newline\\\arraybackslash\hspace{0pt}}m{#1}}
\newcolumntype{C}[1]{>{\centering\let\newline\\\arraybackslash\hspace{0pt}}m{#1}}
\newcolumntype{R}[1]{>{\raggedleft\let\newline\\\arraybackslash\hspace{0pt}}m{#1}}
\usepackage[font=small,skip=5pt]{caption}

\usepackage{booktabs} 
\usepackage{colortbl}

\newcommand{\mc}[3]{\multicolumn{#1}{#2}{#3}}
\definecolor{Gray}{gray}{0.85}
\definecolor{LightCyan}{rgb}{0.88,1,1}

\newcolumntype{a}{>{\columncolor{Gray}}c}
\newcolumntype{b}{>{\columncolor{white}}c}

\definecolor{newcolor}{rgb}{.8,.349,.1}
\newcommand{\revision}[1]{#1} 

\journal{Medical Image Analysis}

\begin{document}

\verso{Shuai Chen \textit{et~al.}}

\begin{frontmatter}

\title{Source Identification: A Self-Supervision Task for Dense Prediction}

\author[1]{Shuai \snm{Chen}\corref{cor1}}
\author[1]{Subhradeep \snm{Kayal}\corref{cor1}}
\author[1,2]{Marleen \snm{de~Bruijne}}

\cortext[cor1]{*: S. Chen and S. Kayal contributed equally.}
\address[1]{Biomedical Imaging Group Rotterdam, Department of Radiology~\&~Nuclear Medicine, Erasmus MC, Rotterdam, The Netherlands.}
\address[2]{Machine Learning Section, Department of Computer Science, University of Copenhagen, DK-2110 Copenhagen, Denmark.}

\received{\today}
\finalform{\today}
\accepted{\today}
\availableonline{\today}
\communicated{Name}

\begin{abstract}
The paradigm of self-supervision focuses on representation learning from raw data without the need of labor-consuming annotations, which is the main bottleneck of current data-driven methods. Self-supervision tasks are often used to pre-train a neural network with a large amount of unlabeled data and extract generic features of the dataset. The learned model is likely to contain useful information which can be transferred to the downstream main task and improve performance compared to random parameter initialization. In this paper, we propose a new self-supervision task called source identification (SI), which is inspired by the classic blind source separation problem. Synthetic images are generated by fusing multiple source images and the network's task is to reconstruct the original images, given the fused images. A proper understanding of the image content is required to successfully solve the task. We validate our method on two medical image segmentation tasks: brain tumor segmentation and white matter hyperintensities segmentation. The results show that the proposed SI task outperforms traditional self-supervision tasks for dense predictions including inpainting, pixel shuffling, intensity shift, and super-resolution. Among variations of the SI task fusing images of different types, fusing images from different patients performs best. 
\end{abstract}


\end{frontmatter}


\section{Introduction}
\label{sec:introduction}

The success of deep learning, and in particular convolutional neural networks (CNNs), may be partially attributed to the exponential increase in the amount of available annotated data. However, in highly specialized domains such as medical image segmentation, it is much harder to acquire precise and dense annotations. Self-supervision is one research direction that enables the network to learn from images themselves without requiring labor-consuming annotations, where the learned features might be useful for the downstream tasks, such as classification and segmentation.

In general, self-supervised learning refers to a collection of approaches that deliberately withhold information in the original data and task a neural network to predict the missing information from the existing incomplete information. In doing so, the network is encouraged to learn general-purpose features which have been found to transfer well to downstream tasks \citep{journals/corr/abs-1902-06162}. The self-supervision pipeline often employs a pre-train and fine-tune strategy. The first step is to pre-train a CNN on a large volume of unannotated samples using a manually designed proxy task, in which the CNN explores and learns generic features of the data itself. The learned features may contain meaningful information of the image data, e.g., intensity distribution, spatial coherence, and anatomical knowledge in medical imaging, etc., depending on how the proxy task is designed. The second step is to fine-tune this pre-trained network on the target (main) downstream task that we are more interested in, which usually has a small set of annotated data in practice. We expect that by exploiting unannotated data and restarting the training from a set of rich pre-trained features, a more robust model on the main task can be trained.

In this paper, we propose a novel self-supervision task called \textit{Source Identification} (SI), which is inspired by the classic Blind Source Separation (BSS) problem. The proposed task is able to train a dense prediction network in a self-supervised manner using unlabeled data. 

\textbf{Contributions:}

1. We propose a novel self-supervision task, SI, wherein a neural network is (pre-)trained to identify one image (source) from mixtures of images. This way, both encoder and decoder are trained and the network is encouraged to learn not only local features but also global semantic features to identify and separate the target source signal. To the best of our knowledge, this is the first BSS-like self-supervised method for deep neural networks.

2. We investigate the ill-posed source identification problem and show in which settings it can be solved by a neural network. The proposed SI method provides a straightforward way to avoid task ambiguity.

3. We conduct extensive experiments on public datasets for two medical image segmentation applications: brain tumor segmentation and white matter hyperintensities segmentation, both from brain MRI. We compare with various existing self-supervision tasks. The results show that the proposed SI method outperforms self-supervision baselines including inpainting, pixel shuffling, intensity shift, and super-resolution in segmentation accuracy in both applications.

\section{Related Work}
\label{sec:relatedwork}

\subsection{Self-Supervision Tasks}
\label{sec:PreviousSST}

Self-supervision is an active research direction in machine learning, permeating from computer vision to natural language processing \citep{mikolov2013efficient,kiros2015skip,devlin2018bert}. In imaging, early self-supervision tasks could be grouped into two main categories: reconstruction based and context prediction based. For example, inpainting is a popular reconstruction based self-supervision task \citep{DBLP:conf/cvpr/PathakKDDE16} where areas in an image are hidden and then reconstructed using a CNN. In a similar fashion, recolorization can be done by removing color of the image and training a CNN to recover it \citep{DBLP:conf/cvpr/LarssonMS17}, and super-resolution by recovering the original resolution of an image from a downsampled image \citep{8099502}. On the other hand, context prediction based tasks make the network learn relationships between parts of an image, such as choosing arbitary tiles in an image and predicting their relative spatial locations \citep{doersch2015unsupervised}. An improved version of this method can be seen in \cite{Noroozi2016UnsupervisedLO}, where tiles were chosen, shuffled and the network was taught to identify the shuffle pattern, thereby forcing it to learn how the tiles make up the original image. Self-supervision has also been applied in medical imaging \citep{10.1007/978-3-030-32251-9_42}, including inpainting \citep{CHEN2019101539,DBLP:conf/miccai/KayalCB20} and puzzle solving by treating a 3D image as a shuffled Rubik's cube \citep{DBLP:conf/miccai/ZhuangLHMYZ19}.
 
All the above self-supervision tasks are designed to learn useful features from a single input image by recovering information withhold from the image itself. However, the rich information that discriminates one image from another is not explicitly considered. The source identification task in this paper aims to learn not only features that can identify each image but also features that can distinguish one image from different images within the dataset. 

Our proposed task shares some similarities with the contemporary contrastive learning method \citep{BeckerHinton92}, which is also gaining popularity in medical imaging \citep{jiao2020self,li2021multi,chaitanya2020contrastive, feng2021nuc2vec,li2020contrastive}. In contrastive learning, the neural network is tasked with recognizing the similarity or dissimilarity of a pair of images input to it, which can be categorized as a context prediction-based rather than reconstruction-based method. As an example, the state-of-art method known as \textit{SimCLR} \citep{49372} works by drawing random samples from the original dataset, applying two augmentations (both sampled from the same family of augmentations) on the samples to create two sets of views. Then these views are passed through a CNN and a fully connected neural network layer to generate latent representations. Finally, these representations are used to train the network, such that the augmented views from the same class are pushed together and the augmented views from different classes are repelled using a contrastive loss. This may encourage the latent features to be more compact and separated, which may provide additional regularization for optimizing the network. However, most contrastive learning approaches are aimed at the downstream task of classification, pretraining only the encoder portion of the network. Thus, in this paper, we focus on the comparison between reconstruction-based methods that are more relevant to our proposed source identification task, as they pre-train the entire network and focus on dense prediction downstream tasks.

\subsection{Blind Source Separation} 

Blind source separation (BSS), also known as signal separation, is the classic problem of identifying a set of source signals from an observed mixed signal. One example of BSS is the cocktail party problem, where a number of people are talking simultaneously in a noisy environment (a cocktail party) and a listener is trying to identify and separate a certain individual source of voice from the discussion. The human brain can handle this sort of auditory source separation problem very well, but it is a non-trivial problem in digital signal processing. Traditional methods such as independent component analysis (ICA) variants are proposed to tackle the BSS problem \citep{bell1995information,amari1996new,hyvarinen2001blind,choi2005blind,isomura2016local}. In the deep learning era, convolutional neural networks have been used to solve BSS problems in signal processing applications such as speech recognition \citep{drude2018deep,drude2019integration} and target instrument separation \citep{chandna2017monoaural}. These works typically employ an encoder network to learn the embeddings of the observed signals and then use traditional techniques like k-means or spectral clustering to cluster the embeddings according to the number of sources. The clustering can also be done by a deep neural network \citep{hershey2016deep}. This paper introduces a BSS-like self-supervised task on image data, in which a neural network is trained that aims to identify and restore the source image content in mixtures with multiple images.

\subsection{Relation to Denoising}

A related task to the proposed source identification is denoising \citep{tian2020deep} which is used to identify and remove undesired imaging artifacts. In denoising, the image and the noise are regarded as two different sources and a model is trained to separate them. The statistical properties of the signal and the noise are very different, unlike in our case, where a mixed image is constructed from images belonging to the same dataset. A denoising network is likely to learn more local features to distinguish noise from clean images rather than high-level semantic features of the image content. Different from the denoising task, the proposed source identification approach tries to separate one image from a fused image with other images rather than with noise. This is a more difficult task that is more likely to capture useful semantic features from the dataset.

\revision{
\subsection{Relation to Mixup}

Mixup was first proposed as a data augmentation strategy while training CNNs in a general setting \citep{zhang2018mixup}, and has been validated to work well in medical image segmentation as well \citep{EatonRosen2018ImprovingDA}. Mixup, in a segmentation setting, works by randomly selecting an image pair from the training data and generating a weighted combination of the input images as well as the target segmentation maps. These generated images are then fed to a CNN during training, in addition to any other data augmentation strategies that may be suitable.

The similarity of our work with Mixup is in the way our mixed images are made, which, in our case, the network learns to identify sources from. However, our approach is a self-supervision strategy, with the aim of teaching the network useful features during pre-training, whereas Mixup is a data augmentation method. Nevertheless, in order to compare the two, we also include a set of experiments with Mixup as an additional data augmentation strategy.}

\section{Methods}
\label{sec:methods}

In Section \ref{sec:def_SI}, we provide a general definition of source identification. In Section \ref{sec:solvable}, we discuss whether and when the source identification task is solvable for a neural network. In Section \ref{sec:source_identification}, we describe how source identification can be used as a proxy task for a self-supervised network. Lastly, we describe four popular competing baseline self-supervision tasks that we compare to in this paper in Section \ref{sec:baseline}. 

\subsection{Definition of The Source Identification Problem}\label{sec:def_SI}

Consider domain $D$, in which each source signal can be distinguished from others, e.g., each signal is an image from a different patient in a medical imaging dataset.
\revision{Multiple ($N$) source signals, $\mathbf{S_N} = (s_1...s_N)^T$, sampled from $D$ are linearly `mixed' to produce $M$ mixtures, $\mathbf{X_M} = (x_1...x_M)^T$, using an $M \times N$ matrix $\mathbf{W}$:
\begin{equation}
\label{equ:generalsi}
\mathbf{X_M} = \mathbf{W}\mathbf{S_N}
\end{equation}
The blind source separation (BSS) problem is to reconstruct individual signals that constitute the mixtures without knowing the transformation $\mathbf{W}$ and the original signals $\mathbf{X}$.}

In the context of employing neural networks for this task, in every training iteration batch we can create $\tilde{M}$ mixtures from $\tilde{N}$ samples, as allowed by the batch-size chosen. Typically, $\tilde{M} <= M$ and $\tilde{N} <= N$. For example, two randomly sampled signals, $s_1$ and $s_2$, can result in a signal mixture, $x$, created by a linear combination:
\begin{equation}
\label{equ:linearcombination}
\begin{split}
&x={w}s_{1}+(1-w)s_{2},\quad{w} \in [0,1]\\
\end{split}
\end{equation}
where the weight, $w$, is a scalar sampled uniformly between 0 and 1. Many mixed signals created in the way above would make up a batch to train the neural network on.

\revision{To learn to separate the signals, we can train a multi-channel neural network model, $f^{\tilde{M}}_{\tilde{N}} (\cdot;\theta)$ parameterized by $\theta$ and with $\tilde{M}$ input and $\tilde{N}$ output channels, to learn the optimal parameters by minimizing the loss, $\mathcal{L}(\theta)$:
\begin{equation}
\label{equ:generalloss}
\mathcal{L}(\theta)=\frac{1}{B}\sum_{\substack{ b \in 1...B}}\ell\left(\mathbf{S}^b_{\tilde{N}}, f^{\tilde{M}}_{\tilde{N}}(\mathbf{X}^b_{\tilde{M}}; \theta)\right)
\end{equation}
where, $B$ is the batch-size, $\mathbf{S}_{\tilde{N}} = (s_1...s_{\tilde{N}})$ is a collection of $\tilde{N}$ randomly picked sources such that $\mathbf{S}^b_{\tilde{N}}$ is the $b^{th}$ collection in the batch, and similarly $\mathbf{X}_{\tilde{M}} = (x_1...x_{\tilde{M}})$ is a collection of `mixtures' created using the process described in Equation \eqref{equ:linearcombination} applied to the sources in $\mathbf{S}^{\tilde{N}}$. In essence, the multi-channel network, $f_{\tilde{M}, \tilde{N}} (\cdot;\theta)$, consumes a single $\mathbf{X}^b_{\tilde{M}}$ as input to produce $\mathbf{S}^b_{\tilde{N}}$ as output, on which the loss is calculated. The loss above is depicted for one batch of one iteration through all of the data available.}

The function $\ell(\cdot,\cdot)$ is composed of the $L_1$ and $L_2$ norm of the difference between the original source signal and the corresponding model output:
\begin{equation}
\label{equ:lossdescription}
\ell(\mathbf{S}_{\tilde{N}}, f^{\tilde{M}}_{\tilde{N}}(\mathbf{X}_{\tilde{M}})) = \frac{1}{\tilde{N}}\sum_{{\substack{ n \in 1...\tilde{N}}}} \left[ \left|s_n - f^{\tilde{M}}_{n}(\mathbf{X}_{\tilde{M}})\right|_1 + \left|\left|s_n - f^{\tilde{M}}_{n}(\mathbf{X}_{\tilde{M}})\right|\right|_2 \right]
\end{equation}
where $f^{\tilde{M}}_{n}(\mathbf{X}_{\tilde{M}})$ is the output image from the $n^{th}$ channel of the neural network acting upon the input $\mathbf{X}_{\tilde{M}}$ described in the previous paragraph.

In the absence of any constraints or assumptions about the properties of the source signals and mixtures, the BSS problem may be ill-posed. For example, if we observe only a few mixtures for the number of source signals to be reconstructed (such that $\tilde{M} <= \tilde{N}$), and/or the overall statistical properties of the signals are not very different, then the mixed signals may not be separable by a network. We show this in the next section with a simple example, and discuss how it motivates our proposed technique.

\subsection{Is Source Identification Solvable?}
\label{sec:solvable}

\begin{figure}
\centering
\includegraphics[width=9cm]{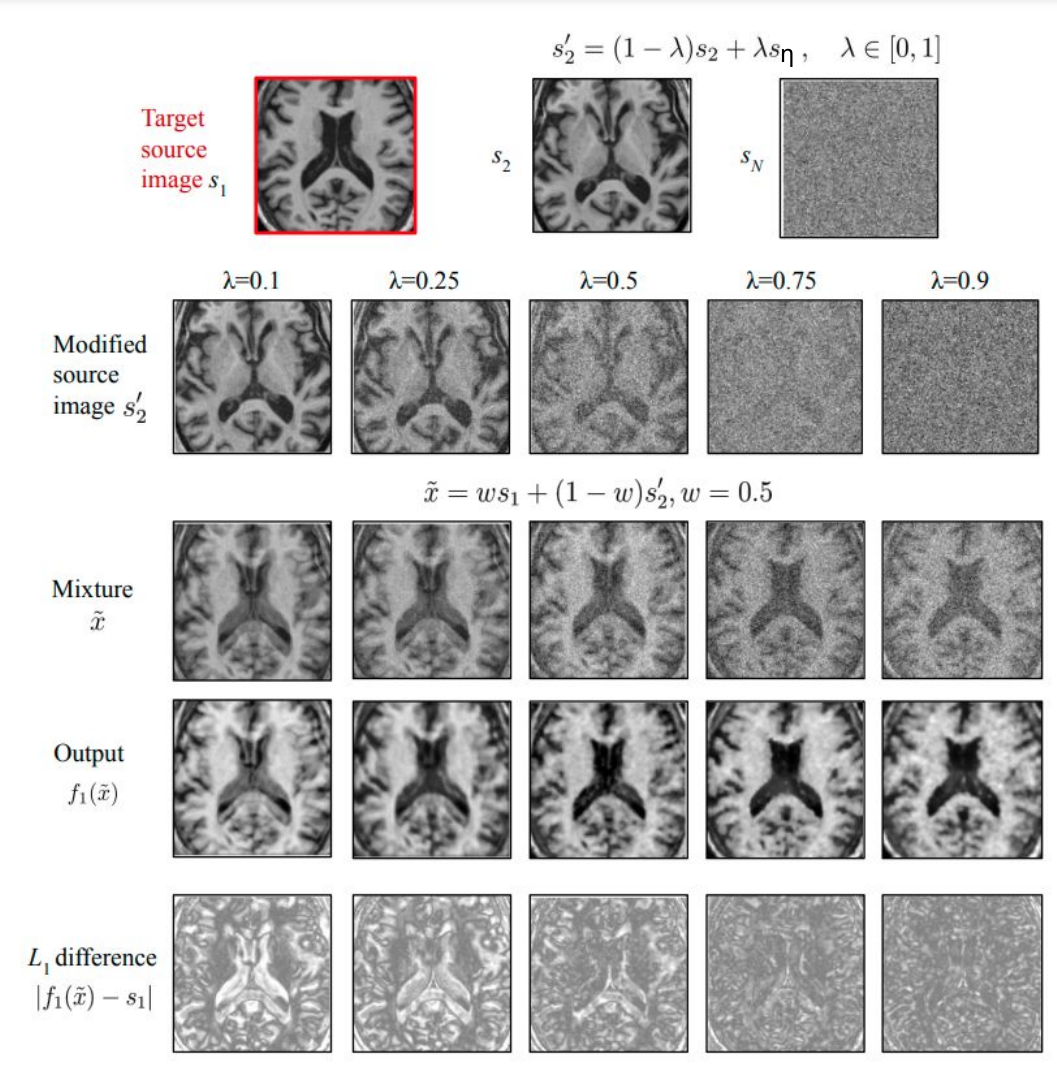}
\caption{\textbf{Qualitative results of recovering $s_1$ with various $\lambda$.} We can see that the model is able to separate and reconstruct $s_1$ from $\tilde{x}$ gradually when $\lambda$ increases from 0.1 to 0.9. The dataset in this experiment contains 30 brain MRI scans from 30 different patients. Best viewed with zoom.}
\label{fig:whensolvable}
\end{figure}

In this small experiment, for a given medical imaging dataset, $D_I$, we randomly sample two source signals (images), $s_1$ and $s_2$, for every training iteration and linearly `mix' them. The mixed signal is fed as input to a neural network while $s_1$ is set as the source to be reconstructed by the network. As can be observed, we have $\tilde{M} = \tilde{N} (=1)$ in this case, and the sources are sampled from the same distribution, which render this ill-posed.

One way to make the reconstruction task less ambiguous would be to sample signal $s_2$ from a different domain than $s_1$, for instance by adding noise to $s_2$, as follows:
\begin{equation}
\label{equ:noisyxb}
s'_2=(1-\lambda){s_2}+\lambda{s_\eta},\quad\lambda \in [0,1]
\end{equation}
where $s_{\eta k}\sim \mathcal{N}(0,1)$ for the $k$th voxel in $s_N$, such that when $\lambda=1$, $s'_2$ is purely a Gaussian noise which belongs to an obviously different domain compared to the imaging domain in dataset $D_I$. The new mixture, $x$, can be created by applying Equation \eqref{equ:linearcombination} on $s_1$ and $s'_2$, and the loss to be minimized by the neural network can be calculated using Equations \eqref{equ:generalloss} and \eqref{equ:lossdescription} by setting $\tilde{M} = \tilde{N} =1$.

The results of a neural network (we use 2D UNet \citep{ronneberger2015u} here) optimized to minimize the reconstruction loss of $s_1, s'_2$ with various value of $\lambda$ are visualized in Figure \ref{fig:whensolvable}. It can be observed that when $\lambda$ is small (0.1), the output is an average of the two images $s_1$ and $s_2$ and the model fails to separate $s_1$ from the mixture,  $x$. When $\lambda$ gradually increases (to 0.9), $s_1$ becomes clearer and better separated.

As this experiment illustrates, the network can not separate sources when they are sampled from the same distribution and mixtures are made arbitrarily. To impose extra constraints and increase separability, one simple way is to sample sources from different domains, for instance an MRI scan and Gaussian noise. However, the case $\lambda=1$ is similar to a self-supervised denoising task where the model may focus on learning the differences between image domain and noise domain. These learned features may contain trivial local patterns and may be less likely to provide useful semantic features for downstream tasks like segmentation. The technique that we propose next relies on creating more mixtures than samples to be extracted, such that $\tilde{M} > \tilde{N}$, by always having a fixed source in all of the mixtures created, such that the network can get extra information to help it identify this desired (fixed) source. We name the former variant as \textit{Denoising SI} (DSI) and the latter variants are \textit{Cross-patients SI} (CSI) and \textit{Within-patients SI} (WSI), depending on which sources are mixed. We describe these variants in detail in the forthcoming sections.

\begin{figure*}
\centering
\includegraphics[width=18cm]{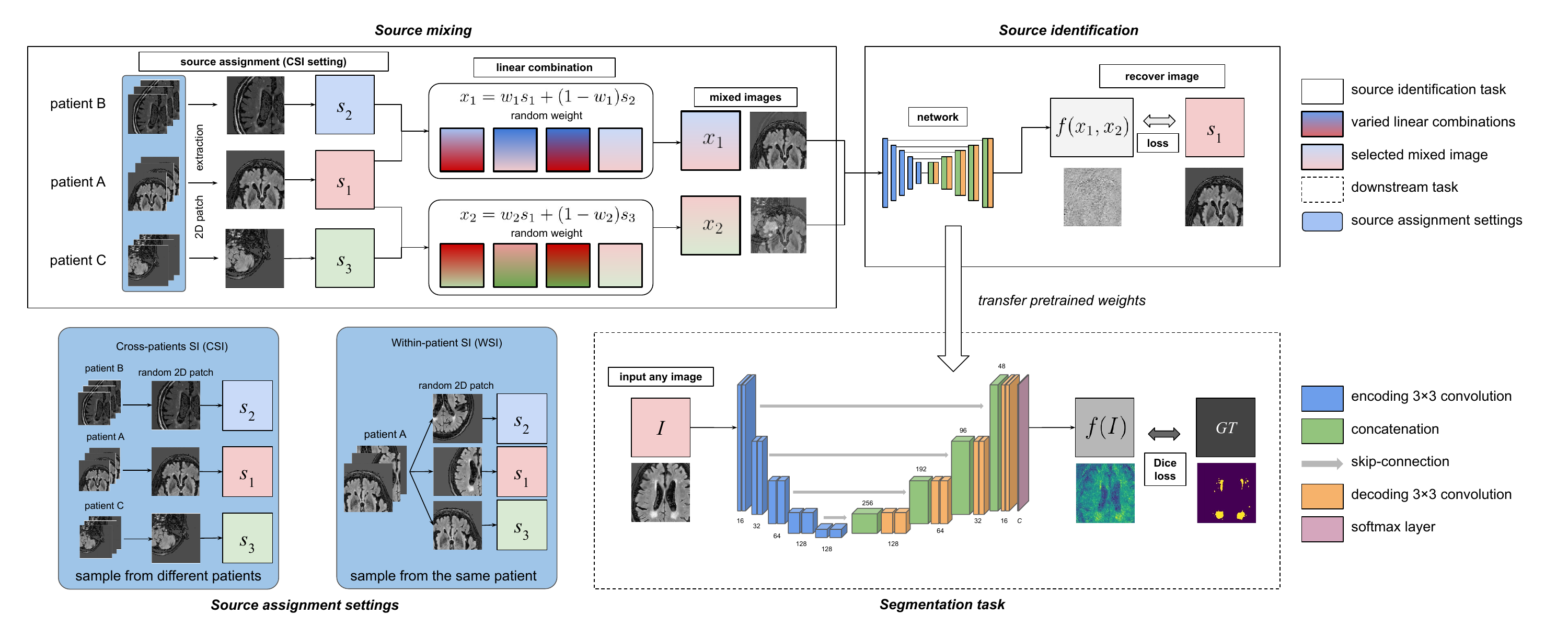}
\caption{\textbf{The proposed source identification task.} Three source images $s_1$, $s_2$, and $s_3$ are used for this illustration. Cross-patients SI (CSI) and within-patient SI (WSI) are two different strategies to extract source signals, which focus on learning features between different patients and within one individual patient respectively. $2\times2$ downsampling and upsampling are applied between different resolutions in the UNet. Best viewed in color with zoom.}
\label{fig:main}
\end{figure*}

\subsection{Proposed Source Identification Task}
\label{sec:source_identification}

In this paper, we propose a simple variation of the source identification task that solves the issue of the ill-posed task. In this task, we sample sources such that one of the sources is present in every input mixture, and make the only target output. This assumes the number of input mixtures, $\tilde{M}$, is set to two or larger. In the case of $\tilde{M}=2$ and $\tilde{N}=1$, the proposed task would be to identify and separate the target signal, e.g., $s_1$, from two mixtures $x_1$ and $x_2$:
\begin{equation}
\label{equ:twomix}
\begin{split}
&x_1={{w}_1}s_{1}+(1-{w}_1)s_{2},\quad{w}_1 \in [0,1]\\
&x_2={{w}_2}s_{1}+(1-{w}_2)s_{3},\quad{w}_2 \in [0,1]\\
\end{split}
\end{equation}
where $w_1$ and $w_2$ are scalars, sampled uniformly between 0 and 1.

Now, the loss function for this arrangement can be written (using Equation \eqref{equ:generalloss}) as:
\begin{equation}
\label{equ:twomixloss}
\mathcal{L}(\theta)=\frac{1}{B}\sum_{{\substack{ b \in 1...B}}}\ell(s^b_1, f^{2}_{1}((x^b_1, x^b_2);\theta)
\end{equation}
where the superscript $b$ denotes samples from a particular batch. The order of input, $(x^b_1,x^b_2)$ and $(x^b_2,x^b_1)$ are equivalent since the mixtures are statistically exchangeable due to the random sampling, and both share the same ground truth, in this case $s^b_1$. It should be noted that even though all source signals are sampled from the same domain $D_I$, this task is solvable for a neural network since the target source signal is specific and invariant, and the number of mixtures is larger than the number of signals to be separated. The workflow of the proposed task is shown in Figure \ref{fig:main}.

It is worth mentioning that although it is trivial to solve the linear equations in Equation \eqref{equ:twomix} and obtain $s_1$, $s_2$, and $s_3$ analytically, it is non-trivial for the network to solve when formulated as a learning problem for a neural network. For example, with the use of data augmentation as well as uniformly sampling the mixing weights while training, the possibility of getting the exact same inputs and outputs is very low, because of which it is unlikely that the network will learn to memorize patterns. \revision{This makes the proposed SI variant an efficient way to learn useful features from a dataset, without labor-consuming annotations, and avoid ambiguity at the same time.} Compared to introducing a different domain to solve the ambiguity problem in Section \ref{sec:solvable}, the proposed method focuses on the same domain which is more likely to learn useful features for the downstream tasks.

\begin{figure}[h]
\centering
\includegraphics[width=7cm]{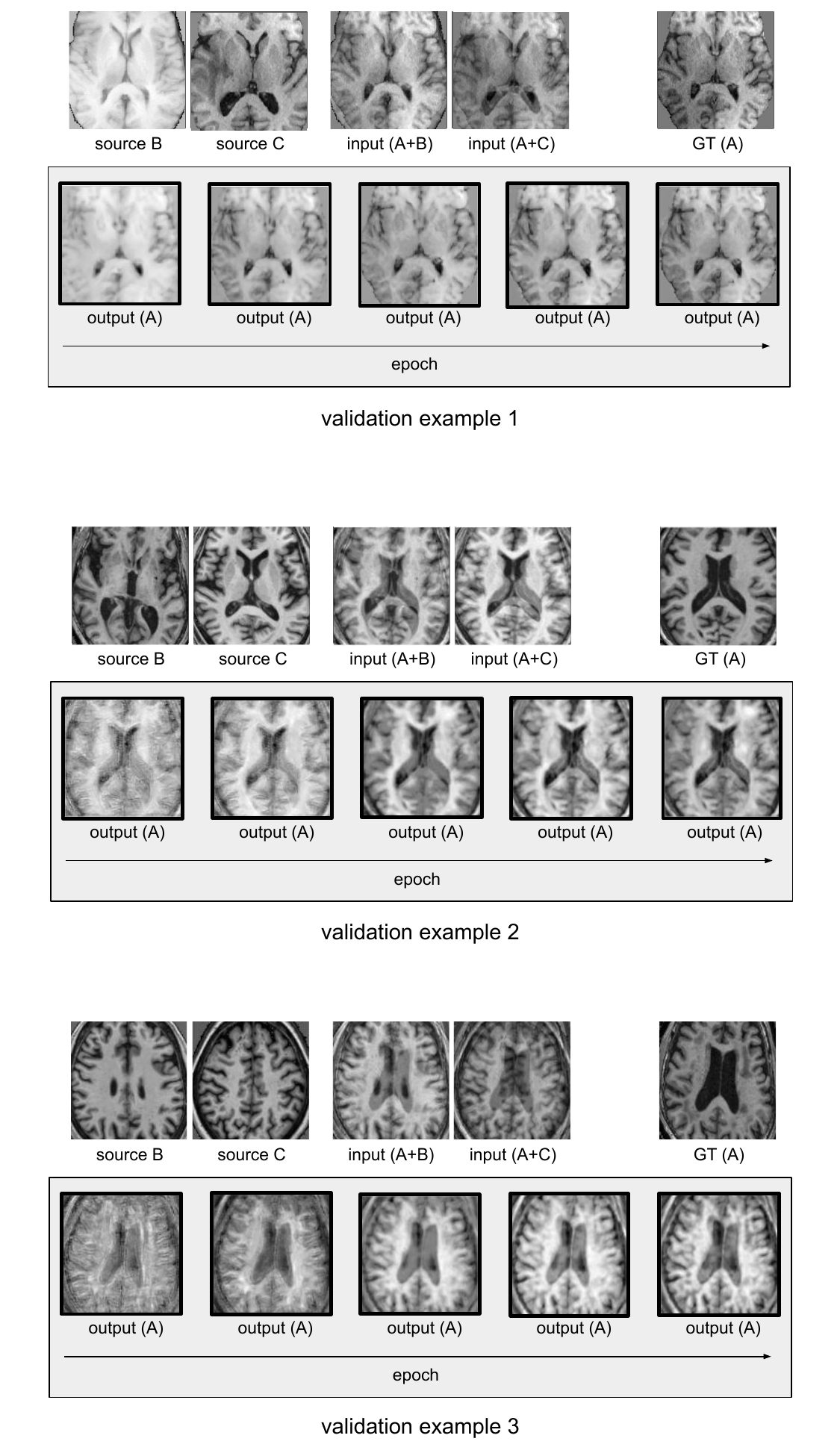}
\caption{\textcolor{black}{\textbf{Visualization of the network solving the SI task while being trained on the BraTS (example 1) and WMH (example 2 and 3) dataset.} The T1 image is used for visualization. The subfigures show three different validation samples consisting of mixed input images, with the second row in each subfigure showing the network output for five intermediate epochs.} The Cross-patients SI (CSI) setting (with three sources) is visualized. The details of the source setting can be seen in Section \ref{sec:sourcesettings}. We can see that the network is able to identify and reconstruct the target source signal $A$ from the input mixtures $A+B$ and $A+C$ gradually during training.}
\label{fig:examples}
\end{figure}

\section{\textcolor{black}{Baseline Self-supervision Tasks}}
\label{sec:baseline}

We compare the proposed method to four widely used self-supervision tasks for dense prediction \citep{shurrab2021selfsupervised, 10.1007/978-3-030-32251-9_42}. The first three tasks focus on the reconstruction and context-based prediction in an image, while the last task focuses on the intensity correction.

\subsection{Inpainting}
Image inpainting is the process of reconstructing the missing or damaged contents of an image, historically employed for restoring paintings and photographs \citep{10.1145/344779.344972}. Inpainting, as a self-supervision task, proceeds by intentionally masking selected areas within an image and a network must learn to recover the missing content. 

In this paper, we implement inpainting self-supervision by overlaying an image $I$ with a regular grid $G$ of a fixed size and randomly masking selected grid cells. Formally, a selected grid cell of pixels, indicated as $g(I)$, where $g\in{G}$, is transformed as:
\begin{equation}
    g'(I) =
  \begin{cases}
    g(I)   & \text{if $\mathbb{B}(\gamma) = 1$}, \\
    0     & \text{otherwise}.
  \end{cases}
\end{equation}
where $\mathbb{B}(\gamma)$ follows a Bernoulli distribution with $\gamma$ probability of being $1$. $\gamma$ is a hyperparameter ranged from 0 to 1. That means in any minibatch, a network only sees approximately $\gamma$ random contents of the input images and tries to predict the rest of them. By masking grids in such a non-deterministic manner, we avoid cases where the network may focus on easy reconstructions and learning trivial features.

The resultant synthetic image $I'$ is made up of all the selected grids, $g'(I)$, thereby retaining $1-\gamma$ fraction of the original image.

\subsection{Local Pixel Shuffling}
Local pixel shuffling has been known to aid a network in learning about the local information within an image, without compromising the global structures \citep{10.1007/978-3-030-32251-9_42}.
This task is similar to inpainting but with additional information on the distribution of intensities to inpaint. In this task, synthetic images are generated by randomly shuffling pixels within the selected grid cell, as shown in the following equation:
\begin{equation}
    g'(I) =
  \begin{cases}
    P g(I) Q   & \text{if $\mathbb{B}(\gamma) = 1$}, \\
    g(I)   & \text{otherwise}.
  \end{cases}
  \label{shufflepix}
\end{equation}
where $\gamma$ is a hyperparameter ranged from 0 to 1 similar to that in inpainting; $P$ and $Q$ are permutation matrices. A permutation matrix is a binary square matrix which can permute the rows of an arbitrary matrix when being pre-multiplied to it, and permute the columns when being post-multiplied. Thus, in the first case of Equation \eqref{shufflepix}, a new grid cell of pixels is generated by shuffling both the rows and columns of the original grid.

\subsection{Super-resolution}
Super-resolution can be implemented as a self-supervision task \citep{zhao2020smore}, wherein a network is trained to deblur the low-resolution image. To create the low-resolution images from high-resolution ones for training, we blur the high-resolution images by transforming every grid cell by replacing all its values with that in the center of the grid:
\begin{equation}
    g'(I) = g(I)_{(\ceil{w/2},\ceil{h/2})}
\end{equation}
where $w$ and $h$ are the width and height of the grid cell $g(I)$. In the training process, given a transformed image as input, the network learns to predict the high resolution version which is the original image before transformation.

\subsection{Non-linear Intensity Shift}
The intensity shift mechanism is proposed by \cite{10.1007/978-3-030-32251-9_42}, where each pixel value in the image is translated monotonically using a Bezier curve (denoted as function $B$) \citep{10.5555/520335}. In medical imaging, since the intensity values in a image usually correspond to the underlying anatomical details, this task can be used to encourage a network to learn useful anatomical features.

Given a voxel value $v$ which is normalized between $[0,1]$, end-points $p_0$, $p_3$, and two control-points $p_1$, $p_2$, the transformed value of the pixel is given by:
\begin{equation}
\begin{split}
v' = B(v) = (1-v^3)p_0 + 3x(1-v^2)p_1 \\
+ 3v^2(1-v)p_2 + v^3p_3
\end{split}
\end{equation}
where points from $p_0$ to $p_3$ are sampled independently at every epoch from a continuous uniform distribution between $0$ to $1$.

\begin{figure*}[t]
\centering
\includegraphics[width=17.5cm]{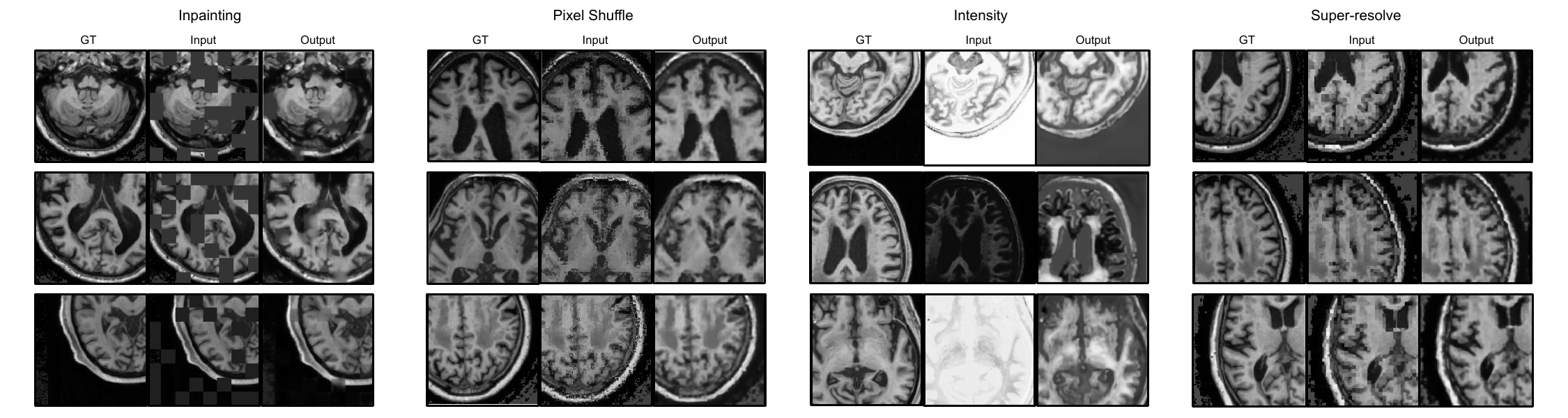}
\caption{\textbf{Visual examples of baseline self-supervised tasks.} The results with best validation performance are used for visualization. Best viewed with zoom.}
\label{fig:visual_ss}
\end{figure*}

\section{Experiments}
\label{sec:experiments}

\subsection{Datasets}
\label{sec:datasets}
We apply our method on two medical imaging segmentation problems: brain tumor segmentation and white matter hyperintensities segmentation. Both datasets contain brain MR images.

\subsubsection{BraTS Dataset}

Multimodal Brain Tumor Segmentation Challenge 2018 \citep{menze2014multimodal,bakas2017advancing,bakas2018identifying} focuses on evaluating methods for the segmentation of brain tumors in multimodal magnetic resonance imaging (MRI) scans. There are in total 210 MR images acquired from different patients. Each MR image contains four modalities: pre-contrast T1-weighted, post-contrast T1-weighted, T2-weighted, and FLAIR. Three brain tumor classes are provided as manual annotations: 1) the necrotic and the non-enhancing tumor core (NCR\&NET); 2) the peritumoral edema (ED); and 3) the enhancing tumor (ET). Since the evaluation classes of the challenge are the combined classes: whole tumor (NCR\&NET+ED+ET), tumor core (NCR\&NET+ET), and enhancing tumor (ET), we use these combined classes for the actual training. We randomly split the dataset in 1) 100 subjects for training the self-supervision tasks and the main segmentation task; 2) 10 subjects for validation; and 3) 100 subjects for testing. For each subject, we cropped/padded MR images into a constant size of $200\times{200}\times{Z}$ ($Z$ is the number of axial slices of the image) where the main brain tissues are preserved. Following the preprocessing of nnUNet \citep{DBLP:journals/corr/abs-1809-10486}, Gaussian normalization (subtracting mean and dividing by standard deviation) is applied on the brain foreground for each modality for each image individually.

\subsubsection{WMH Dataset}

The White Matter Hyperintensities (WMH) Segmentation Challenge \citep{Kuijf} evaluates methods for the automatic segmentation of WMH in brain MR images. The provided MR images contain T1-weighted and FLAIR MR sequences and are acquired from 60 patients, where each group of 20 patients are from a different hospital. The manual segmentation of WMH lesions are also provided for each image. We randomly split the dataset in 1) 30 subjects for training the self-supervision tasks and the main segmentation task; 2) 10 subjects for validation; and 3) 20 subjects for testing. For each subject, we centre-cropped/padded MR images into a constant size of $200\times{200}\times{Z}$, where Z is the number of axial slices in the 3D image. \revision{The cropping/padding of was necessary as images from the different hospitals have slightly different sizes and it was convenient to have images of a constant size to have all of them processed in the same way by the network. Additionally, the size of $200\times{200}$ covers the main brain tissue, which is what the network needs to consume for learning.}
We use Gaussian normalization to normalize the intensities inside the brain foreground similar to the BraTS dataset.

\subsection{Settings for The Proposed SI Task}
\label{sec:sourcesettings}

There are two hyperparameters to adjust in the proposed task. The first one is the process of creating the `mixed' images. In Section \ref{sec:source_identification}, we considered the example where linear combinations of three signals $s_1$, $s_2$, and $s_3$ to create two mixed images. This can be generalized to choose $\tilde{N}$ images to create $\tilde{M}$ mixtures using uniformly sampled weights, such that for every mixture the sampled weights sum to 1. For example, when $N = 5, \tilde{N} = 3, \tilde{M} = 2$:
\begin{equation}
\label{equ:matrixW}
\begin{split}
&x_1={w}_1{s_1}+{w}_2{s_2}+{w}_3{s_3}, \quad{w}_1+{w}_2+{w}_3=1\\
&x_2={w}_4{s_1}+{w}_5{s_4}+{w}_6{s_5}, \quad{w}_4+{w}_5+{w}_6=1
\end{split}
\end{equation}
where all weights are scalars randomly sampled between 0 and 1 under the conditions, and the common image $s_1$ is to be reconstructed. Recall that $N$ is the total pool of randomly chosen source images for the creation of mixed images in a single batch of a training iteration, while $\tilde{N}$ is the number of component images mixed, which is a subset of $N$.

The second hyperparameter is the source assignment strategy. In this paper, we consider three types of source assignment strategies: 

\subsubsection{Cross-patients SI} To make the network learn to identify the target source image and discriminate it from the other source images, $N$ random patients are used to extract signals (2D slice per patient) respectively in every training sample. We refer to this SI variant as \textit{Cross-patients SI} (CSI).

\subsubsection{Within-patient SI}
To make the network focus on each particular source in the dataset, we use the same patient \revision{image} to extract all $N$ signals (all 2D slices from the same patient). Since information from only 1 patient is used in each mixture, the network is unlikely to learn the cross-sources information among different patients. We refer to this SI variant as \textit{Within-patient SI} (WSI).

\subsubsection{Denoising SI}
To investigate the difference between the proposed SI task and the traditional denoising task, we replace sources $s_2$ and $s_3$ in CSI with random Gaussian noise with zero mean and unit variance. This task is similar to a traditional denoising task and would encourage the network to learn representative features that distinguish differently distributed sources like image and noise explicitly. We refer to this SI variant as \textit{Denoising SI} (DSI).

All experiments in Section \ref{sec:experiments} apply the linear combination of constituting three signals in two mixtures as showed in Equation \eqref{equ:matrixW}, where in total $N=5$ signals are used to generate a training sample. This setting is tuned on the validation set for both datasets. To avoid the network from learning trivial features, all combinations without any overlapping between the brain regions are excluded as training samples.

\begin{table*}
  \caption{\textbf{Results of fully-supervised setting.} Each experiment is repeated 3 times with different random data split. For BraTS, the same 100 images are used for training the main segmentation task (with labels) and the self-supervision proxy tasks (without labels); for WMH, the same 30 images used for both labeled and unlabeled data. Mean Dice score (standard deviation) over all experiment testing data is reported for each class individually, where WT=whole tumor, TC=tumor core, ET=enhancing tumor, \textit{All}=WT+TC+ET (only for BraTS), WMH=white matter hyperintensities. \textcolor{black}{*}: significantly better than the CNN baseline ($p<0.05$). \textcolor{black}{$\diamond$}: significantly worse than the CNN baseline ($p<0.05$). P-values are calculated by two-sided paired t-test in each class. Boldface: best and not significantly different from the best results.}
  \label{table:resultsBraTS}
  \small
  \centering
  \begin{tabular}{llllll}
  \toprule
    \mc{1}{b}{\textit{~}}&\mc{4}{a}{BraTS} & \mc{1}{b}{\textit{~}}\\
    Methods/Class & \mc{1}{b}{WT} & \mc{1}{b}{TC} & \mc{1}{b}{ET} & \mc{1}{b}{\textit{All}} & \mc{1}{b}{WMH} \\
    \hline

    CNN & 0.866(0.11) & \textbf{0.835}(0.17) & 0.785(0.16) &  0.846(0.11)\rule{0pt}{2.6ex} & 0.775(0.11)\\
    CNN-restart & 0.868(0.11) & 0.825(0.19)\textcolor{black}{$\diamond$} & 0.786(0.16) &  0.848(0.11) & 0.781(0.11)\\
    Inpainting & 0.867(0.11) & \textbf{0.838}(0.17) & 0.788(0.16) & 0.850(0.11) & 0.782(0.13)\\

    Pixel Shuffle & 0.859(0.13) & 0.829(0.20)\textcolor{black}{$\diamond$} & 0.777(0.18)\textcolor{black}{$\diamond$} & 0.844(0.13) & 0.777(0.12)\\
    
    Intensity & 0.865(0.12) & \textbf{0.838}(0.18) & 0.787(0.16) & 0.846(0.12) & 0.775(0.13)\\
    
    Super-resolve & 0.852(0.13)\textcolor{black}{$\diamond$} & \textbf{0.838}(0.18) & 0.786(0.17) & 0.842(0.13) & 0.776(0.12)\\
    
    \hline
    Denoising SI & 0.868(0.09)& 0.821(0.17)\textcolor{black}{$\diamond$}  & 0.783(0.16) & 0.850(0.10) & 0.771(0.12) \rule{0pt}{2.6ex}\\
    Within-patient SI & 0.869(0.09) & 0.817(0.19)\textcolor{black}{$\diamond$} & 0.781(0.16) & 0.851(0.10) & 0.769(0.12) \\
    Cross-patients SI (ours) & \textbf{0.878}(0.09)\textcolor{black}{*} & \textbf{0.837}(0.17) & \textbf{0.796}(0.15)\textcolor{black}{*} & \textbf{0.861}(0.09)\textcolor{black}{*} & \textbf{0.793}(0.11)\textcolor{black}{*}\\
    \bottomrule
  \end{tabular}
\end{table*}

\subsection{Settings for The Baseline Tasks}
For inpainting, the grid size is tuned from ranging $[2,2]$ to $[64,64]$ and the masking percentage ranging from 0\% to 100\%; for local pixel shuffling and super-resolution, the grid size has the same tuning range as that in inpainting. There is no hyperparameter to tune for non-linear intensity shift. All hyperparameters are tuned on the validation set on the main task performance.

\subsection{Network Architecture}

We use the same network backbone for both the self-supervision proxy tasks and the segmentation main task. It is based on 2D UNet \citep{ronneberger2015u}  and details of the network are shown in Figure \ref{fig:main}. The network has two input-output layer settings: 1) for training the proposed SI task, the input layer has $T\times{2}$ channels where $T$ is the number of imaging modalities for the two input mixtures. The output layer has $T$ channels for reconstructing all modalities of $s_1$; 2) for segmentation, the input layer is replaced with a new layer with $T$ channels for the input image $x$ and the output layer is replaced with a layer with $C$ channels for the segmentation predictions where $C$ is the number of classes. All the intermediate layers are shared between the pretrained proxy task and the main task. When no pretrained network is used, the weights of all convolutional layers are initialized by Kaiming initialization \citep{he2015deep}.

\revision{
The choice of the network parameters are influenced by the state-of-the-art nnUNet \citep{DBLP:journals/corr/abs-1809-10486} model, described in a Section \ref{sec:dataaugment}.}

\subsection{Training Strategy and Data Augmentation}
\label{sec:strategy}

We conduct main experiments in a fully-supervised setting and a semi-supervised setting for both datasets. 

\subsubsection{Fully-supervised Setting} There are two steps to train the network in a self-supervised manner. First, we need to pre-train the network with the corresponding proxy task, \revision{as described in Sections \ref{sec:source_identification} and \ref{sec:baseline}. The proxy task uses the same dataset as the main task; e.g., for the BraTS dataset, we pre-train and fine-tune the network on same 100 (labeled) images from the training set.} A batchsize of 1 is used for the proxy task for all experiments in this paper, \revision{realized via tuning from 1 to 4, based on the validation set. Next, for the main task, we use a batchsize of 8 and 4 for training the main task in BraTS and WMH, respectively, obtained by tuning between 1 to 16 based on the validation set}.

\subsubsection{Semi-supervised Setting} \revision{In the fully supervised setting, we utilize the entire training dataset to pre-train and fine-tune the network. Since the strength of self-supervision comes from a network needing a much smaller volume of data to be fine-tuned, we also conduct experiments to test this hypothesis, which we call the semi-supervised setting. In this setting, the network is pre-trained on the entire training dataset but fine-tuned on only a fraction of the training data. 25 of 100 labeled images were used from the training set to fine-tune the pre-trained model for BraTS; for WMH we used only 5 images.} The same batchsize is used for both proxy task and main task as was used in the fully-supervised setting.

\subsubsection{Data Augmentation and Optimization Parameters}
\label{sec:dataaugment}

Random rotation, scaling, flipping, and elastic deformation are applied to the original 2D images as data augmentation for all experiments. Following the nnUNet paper, we use SGD optimizer and 'poly' learning rate policy ($1-(\text{epoch}/\text{epoch}_{max})^{0.9}$), where $\text{epoch}_{max}=1000$ and for the BraTS dataset and 10000 for WMH, with the initial learning rate $1\times{10^{-2}}$, momentum $0.99$, and weight decay $3\times{10^{-5}}$ for both the proxy task and the main task. Early stopping is applied when there is no improvement for 50 epochs to avoid overfitting to the validation set. We also tried restarting the optimization for the main task after optimizing from random initialization, \revision{which we call \textit{CNN-restart}} for both datasets for fair comparison.

\section{Results}
\label{sec:results}

\subsection{Segmentation Results}
\label{sec:segresults}

Table \ref{table:resultsBraTS} shows the segmentation results for the two datasets in the fully-supervised setting. The proposed Cross-patients SI method achieves the best average performance (except TC: the tumor core in BraTS) in both datasets and shows significant improvement over the other baselines and SI variants in four out of five classes (WT: whole tumor, ET: enhancing tumor, \textit{All}: WT+TC+ET, and WMH). The \textit{All} class calculates the Dice coefficient of WT+TC+ET together (by concatenating the three classes but not summing up them into one class) and is the most important one in BraTS.  

Among the three different settings of source identification task (CSI, WSI, and DSI), CSI achieves the best results with a Dice score of 0.861 (All) and 0.793 in BraTS and WMH datasets separately, which is significantly better than WSI and DSI. WSI and DSI have similar performance in both datasets and are not significantly different from each other. This suggests the importance of the cross-source setting. One reason could be that compared to WSI and DSI, CSI is using the data more efficiently where the network sees more source images per epoch. It should also be noted that the pixel shuffle task shows worse performance than the CNN baseline in four out of five classes (significant in TC and ET classes). In the tumor core (TC) segmentation, four methods (inpainting, intensity shift, super-resolve, and CSI) show comparable improvements to the CNN baseline (not significant to each other), which indicates the efficiency of different self-supervised methods may vary through different classes and the tumor core segmentation is more difficult to improve compared to other classes. Nevertheless, overall, the proposed CSI can provide a better starting point for the segmentation task than most of the self-supervision baseline tasks.

\subsection{Semi-supervised Results}
\label{sec:semi}

\begin{table*}
  \caption{\textbf{Results of semi-supervised setting.} The best results are marked in bold. Each experiment is repeated 3 times with different random data split. For BraTS, all 100 training images are used to train the unlabeled self-supervised task; fine-tuning is performed on 25 of the training images using the segmentation labels and the 25 labeled images are contained in the 100 unlabeled images. For WMH, 5 images are used for labeled data and 30 images are used for unlabeled data and the 5 labeled images are contained in the 30 unlabeled images. Mean Dice score (standard deviation) over all experiment testing data is reported for each class individually, where WT=whole tumor, TC=tumor core, ET=enhancing tumor, \textit{All}=WT+TC+ET (only for BraTS), WMH=white matter hyperintensities. \textcolor{black}{*}: significantly better than the CNN baseline ($p<0.05$). \textcolor{black}{$\diamond$}: significantly worse than the CNN baseline ($p<0.05$). P-values are calculated by two-sided paired t-test in each class. Boldface: best and not significantly different from the best results.}
  \label{table:resultsBraTSsemi}
  \small
  \centering
  \begin{tabular}{llllll}
  \toprule
    \mc{1}{b}{\textit{~}}&\mc{4}{a}{BraTS} & \mc{1}{b}{\textit{~}}\\
    Methods/Class & \mc{1}{b}{WT} & \mc{1}{b}{TC} & \mc{1}{b}{ET} & \mc{1}{b}{\textit{All}} & \mc{1}{b}{WMH} \\
    \hline

    CNN & 0.823(0.11) & 0.780(0.21) & 0.743(0.19) & 0.816(0.12)\rule{0pt}{2.6ex} & 0.739(0.16)\\
    
    CNN-restart & 0.821(0.13) & 0.775(0.22) & 0.739(0.19) &  0.812(0.13) & 0.731(0.16)\textcolor{black}{$\diamond$}\\
    
    Inpainting & 0.842(0.15) & \textbf{0.817}(0.20)\textcolor{black}{*} & 0.754(0.18)\textcolor{black}{*} & 0.827(0.15) & \textbf{0.761}(0.12)\textcolor{black}{*}\\

    Pixel Shuffle & 0.823(0.17) & 0.782(0.23) & 0.723(0.21)\textcolor{black}{$\diamond$} & 0.806(0.17) & 0.744(0.15)\\
    
    Intensity & 0.832(0.16) & 0.804(0.21)\textcolor{black}{*} & 0.746(0.19) & 0.817(0.16) & 0.740(0.15)\\
    
    Super-resolve & 0.848(0.15) & \textbf{0.819}(0.20)\textcolor{black}{*} & \textbf{0.760}(0.19)\textcolor{black}{*} & 0.829(0.14) & 0.756(0.13)\textcolor{black}{*}\\
    
    \hline
    Denoising SI & 0.823(0.16)\textcolor{black}{$\diamond$} & 0.776(0.22)  & 0.747(0.20) & 0.804(0.16)\textcolor{black}{$\diamond$} & 0.755(0.13)\rule{0pt}{2.6ex}\\
    Within-patient SI & 0.836(0.13) & 0.779(0.20) & 0.749(0.18) & 0.814(0.13) & 0.754(0.12)\\
    Cross-patients SI (ours) & \textbf{0.855}(0.12)\textcolor{black}{*} & \textbf{0.811}(0.18)\textcolor{black}{*} & \textbf{0.764}(0.17)\textcolor{black}{*} & \textbf{0.837}(0.12)\textcolor{black}{*} & \textbf{0.783}(0.11)\textcolor{black}{*}\\
    \bottomrule
  \end{tabular}
\end{table*}

We conduct experiments on both datasets in semi-supervised settings in order to investigate how much the proposed self-supervision task would help when only a smaller amount of labeled data is available to train the proxy task. The results are shown in Table \ref{table:resultsBraTSsemi}. Similar trends can be observed from these semi-supervised results compared to those in fully-supervised results. Similar to Table \ref{table:resultsBraTS}, the proposed CSI method gets the largest improvements in BraTS (except the tumor core) and WMH. The improvements are significant compared to all other methods in whole tumor and \textit{All} in BraTS. In WMH, both the proposed CSI method and inpainting are significantly better than the other methods. It should also be noted that when only few labeled images are available, more self-supervision methods show significant improvements compared to CNN baseline (12$\textcolor{black}{*}$ results in Table \ref{table:resultsBraTSsemi} compared to 4$\textcolor{black}{*}$ results in Table \ref{table:resultsBraTS}). This shows the general advantages of feature learning in self-supervision methods compared to CNN baseline. 

The SI variants WSI and DSI still show close performance to each other in most classes and perform significantly worse than CSI. Similar to the fully-supervised setting, the pixel shuffle task does not show improvements compared to the CNN baseline in most classes. It should be noted that the CSI performance in semi-supervised setting (0.837 in BraTS and 0.783 in WMH) is very comparable to the fully-supervised CNN baseline result (0.846 in BraTS and 0.775 in WMH), which required 4 times more training images. Inpainting and super-resolve show better performance than CNN baseline, but still worse than CSI (significant in BraTS). The proposed method shows larger performance improvements in WMH dataset where far fewer labeled data are used compared to BraTS dataset (5 labeled vs. 25 labeled and with 4.4\% vs. 3.1\% Dice improvements to the CNN baseline). This shows in a practical situation in medical imaging where segmentation labels are scarce, a well-designed self-supervision task can still preserve considerable performance given enough unlabeled data.

\subsection{Influence of The Number of Sources}
\label{sec:absources}

\begin{figure}[h]
\centering
\includegraphics[width=7cm]{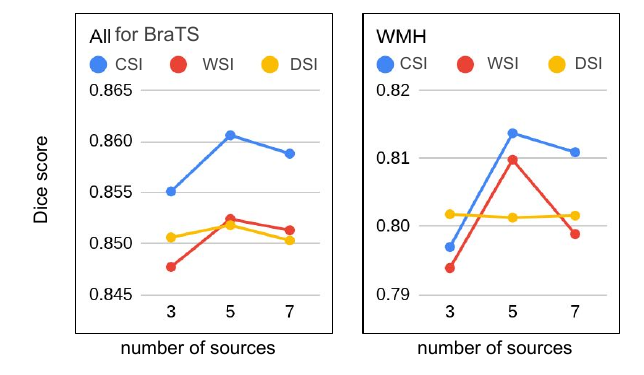}
\caption{\textbf{Influence of the number of fused sources.} The results are obtained by an independent run on BraTS and WMH dataset using the same data in a random data split with fully-supervised setting, similar to Table \ref{table:resultsBraTS}. The number of sources 5 is used for experiments in Table \ref{table:resultsBraTS} and \ref{table:resultsBraTSsemi}. Best viewed in color.}
\label{fig:absources}
\end{figure}

We conduct experiments to investigate the influence of the number of images used in the proposed SI task. $N = 3, 5, 7$ and $\tilde{N} = 2, 3, 4$ sources (e.g. in Equation \eqref{equ:matrixW}, $N = 5, \tilde{N} = 3$) are tested to generate $\tilde{M} = 2$ fused images as input to the network. The experiments are independent runs on BraTS and WMH dataset in fully-supervised setting. Note that the hyperparameters $N$ and $\tilde{N}$ is tuned on the validation set for all experiments. The results are shown in Figure \ref{fig:absources}. We can see that the setting $N = 5, \tilde{N} = 3$ sources setting achieves the best performance in the main segmentation task for CSI and WSI while for DSI the effect is much smaller. Too few sources may make it too easy to reconstruct the target signal which may result in trivial features, while too many sources may make it too difficult to recognize the target, resulting in arbitrary features. 

\revision{
\subsection{Comparison to Mixup}

\begin{table}
  \caption{\textbf{Comparison of Cross-patients SI to Mixup in fully supervised setting.} The best results are marked in bold. Each experiment is repeated 3 times with random data splits, and the mean Dice scores are reported. \textcolor{black}{*}: significantly better than the CNN + mixup ($p<0.05$). Boldface: Highest mean Dice result.}
  \label{table:mixup}
  \small
  \centering
  \begin{tabular}{lll}
  \toprule
    \mc{1}{b}{Methods/Class} & \mc{1}{b}{BraTS WT} & \mc{1}{b}{WMH} \\
    \hline

    CNN & 0.866(0.11)\rule{0pt}{2.6ex} & 0.774(0.11)\\
    CNN + mixup & 0.875(0.11)\rule{0pt}{2.6ex} & 0.801(0.11)\\
    Cross-patients SI & 0.878(0.09)\rule{0pt}{2.6ex} & 0.793(0.11)\\
    Cross-patients SI + mixup & \textbf{0.886}(0.10)*\rule{0pt}{2.6ex} & \textbf{0.803}(0.12)\\

    \bottomrule
  \end{tabular}
\end{table}

Table \ref{table:mixup} shows the results to compare our proposed approach to Mixup in the fully-supervised setting. Here, the Cross-patients Source Identification (CSI) based self-supervised pre-training is compared to the baseline CNN without any pre-training, with and without Mixup as an additional data augmentation strategy. Results show that Mixup improves both the baseline and our proposed approach, with a higher relative improvement for detecting tumours in the BraTS dataset.
}

\section{Discussion}

In this paper, we propose a new self-supervision task named source identification (SI) which is inspired by the blind source separation problem, and we investigate the task ambiguity in the SI problem for neural networks. Unlike most previous reconstruction-based self-supervision tasks that focus on restoring image contents from only one source image, the proposed task enables the network to see multiple images from mixtures and learn to separate the source image from the others and reconstruct it. The experiments show that the proposed method outperforms baseline methods in both datasets including the CNN baseline, restart-CNN with the initial learning rate, and commonly used self-supervised methods inpainting, pixel shuffle, intensity shift, super-resolution, and denoising. The proposed method shows the largest improvements in the  semi-supervised setting when very few labeled data and many unlabeled data are available, which is a common scenario in medical imaging applications.

\subsection{Comparison to Other Self-supervision Methods}

One main difference between the proposed SI task and existing reconstruction-based self-supervision tasks is that SI learns features from not only the remaining part of the same distorted image but also from other images of the same domain. By distinguishing each image from others, potentially useful discriminative features can be learned while reconstructing the target image. These features may better capture general domain knowledge, e.g. anatomy and pathology knowledge, by seeing and comparing different patients' images at the same time. A proper understanding of anatomy and pathology across different individuals is required to successfully solve a single image identification and reconstruction. Features learned by SI may therefore provide a better starting point for optimization of the downstream task than the features learned by previous self-supervision tasks such as inpainting, pixel shuffling, intensity shift, super-resolution, and denoising. 

In this paper, we focus on the comparison between reconstruction-based self-supervised methods, which all use the synthetic distorted image as input and the original target image as ground truth. We consider the context prediction-based methods such as tiles location prediction \citep{doersch2015unsupervised}, puzzle solving \citep{DBLP:conf/miccai/ZhuangLHMYZ19}, contrastive learning \citep{BeckerHinton92} as another category of self-supervised tasks. These methods optimize a predefined classification/regression task based on the information within a single image \citep{doersch2015unsupervised,DBLP:conf/miccai/ZhuangLHMYZ19} or across different images \citep{BeckerHinton92}, and thus they usually do not train a relevant (dense) decoder. On the contrary, the reconstruction-based methods inherently require a dense decoder for learning concrete and high-resolution features and outputting dense pixelwise predictions, which may result in a model that fits better to dense prediction tasks like segmentation.

\subsection{Apply SI using Unlabeled Data with Less Overfitting}
\label{sec:semidiscussion}

Self-supervised learning allows using unlabeled data without additional annotations from experts and pretraining with both labeled and unlabeled data before fully-supervised learning. The quality of the learned features from self-supervised tasks is usually evaluated on downstream tasks like segmentation. In our experiments, larger improvements are observed in the semi-supervised setting compared to fully-supervised setting, especially for the WMH dataset. Our results show that given the same amount of unlabeled data, the proposed SI can learn more useful features from unlabeled data compared to other self-supervised tasks. One reason could be that the proposed SI task may suffer less from the overfitting problem compared to traditional methods like inpainting and super-resolution. For example, given the unlabeled data, the model may try to solve the inpainting or super-resolution task by memorizing the input images and restoring the missing content when there is enough model capacity, which may result in learning trivial features. In contrast, the SI task takes inputs from many more different combinations of images given the same amount of unlabeled data (when $N = 5$ in 100 images, the number of possible image combinations would be the binomial coefficient $C(100,5)\times{5}\approx{3.8\times{10^8}}$), which makes the model more difficult to memorize and overfit to a particular image but has to find a more general way to solve the SI task, e.g. learning anatomy knowledge, which can be non-trivial and useful for downstream tasks like segmentation.

\subsection{Application to Other Dense Prediction Tasks}
\label{sec:applications}

In this paper, we apply the proposed SI method to segmentation, a dense prediction task. The pretrained SI features can also be transferred to other medical imaging dense prediction tasks such as for instance depth estimation \citep{liu2019dense}, image registration \citep{balakrishnan2018unsupervised}, and detection based on distance maps \citep{10.1007/978-3-030-32251-9_26}. Moreover, these tasks may also benefit from the cross-sources features learned in the SI method. For example, a good image registration model may require not only the alignments between local patterns across different modalities (within one patient) but also the general anatomy knowledge across different patients to constrain possible transformations. With a proper design of the proxy dataset and the SI setting, the potential scenarios to apply the proposed method can be greatly extended.

\subsection{Limitations}

It has been studied in literature that the performance of self-supervised approaches differ significantly based on the difficulty of the pretraining task and it's relatedness to the main task \citep{su2020does, you2020does, DBLP:conf/miccai/KayalCB20}. For example, the performance of inpainting as a self-supervision task would suffer when the size of the masked area is too large or too small. Too large the masked area, and the pretraining task would be too difficult to solve; too small and it would be very easy. This would affect the quality of the learned features, and hence the efficiency of the network on the main task. Similarly, for our approach, the performance of the network is determined by how separable the mixed images are and how much information the network needs to learn to separate them.

\begin{figure}
\centering
\includegraphics[width=7cm]{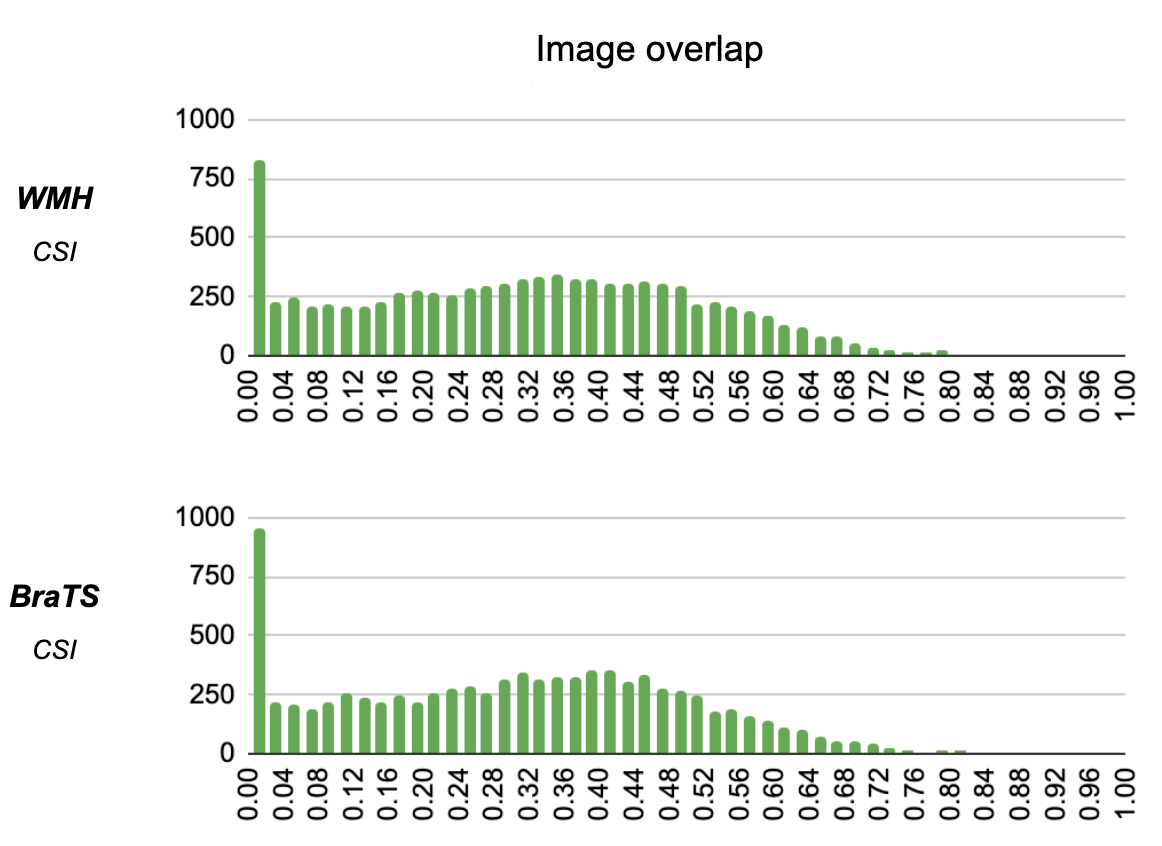}
\caption{\textbf{Distribution of the degree of mixing randomly sampled images.} We observe that this distribution is almost uniform in the two datasets used.}
\label{fig:distributionjs}
\end{figure}

We indirectly test the former hypothesis in Section \ref{sec:solvable}, where it is shown that very similar images would be extremely hard to separate. To explore whether this is a practical problem in our case, we devise a simple statistical experiment. First, pairs of 2D slices are randomly sampled from different images in one dataset (BraTS or WMH), and data augmentation is applied to them as described in Section \ref{sec:dataaugment}. Next, the brain mask is extracted from the resulting images, via simple intensity based thresholding, and the overlap of the corresponding brain masks are measured using Jaccard Similarity. Finally, the distribution of the similarities measured is plotted and shown in Figure \ref{fig:distributionjs}. As we can observe, the similarities almost uniformly range from very low (nearly 0) to moderately high (0.75), indicating that for our datasets, the network would receive a wide range of mixed images for training. As mentioned in Section \ref{sec:sourcesettings}, we exclude all mixed images with 0 similarities (no overlap at all) to avoid the network from learning trivial features. Thus, for our experiments, we do not need extra control of the degree of mixing images.

The second hypothesis revolves around how much information the network needs to learn to identify the source from the mixed images. In Section \ref{sec:absources} we empirically demonstrate the effect of the number of fused sources on the final performance. It is noticed that too few or too many fused sources are detrimental to the efficiency of the network. 

Our proposed approach is sensitive to these two degrees of freedom and, although we have enough empirical evidence for the datasets in question, further testing is required to make a general comment about the sensitivity of our method to these two factors.

\section{Conclusion}

We propose a novel self-supervision task called source identification which is inspired by the classic blind source separation problem. The proposed task is to identify and separate a target source image from mixtures with other images in the dataset, which requires features that are also relevant for the downstream task of segmentation. On two brain MRI segmentation tasks, the proposed method provides a significantly better pretrained model for segmentation compared to other self-supervision baselines including inpainting, local pixel shuffling, non-linear intensity shift, and super-resolution in both fully-supervised and semi-supervised settings. The proposed method can be generalized to other dense prediction applications.

\section*{Acknowledgment}
The authors would like to thank Gerda Bortsova and Hoel Kervadec for their constructive suggestions for the paper, and organizers of BraTS 2018 and WMH 2017 Challenges for providing the public datasets. This work was partially funded by Chinese Scholarship Council (File No.201706170040).

\bibliographystyle{model2-names.bst}\biboptions{authoryear}
\bibliography{refs}

\end{document}